\documentclass{article}
\pdfoutput=1
\usepackage{verbatim}
\usepackage{microtype}
\usepackage{graphicx}
\usepackage{subfigure}
\usepackage{booktabs} 
\usepackage{amsmath,amssymb,amsfonts}
\usepackage{xfrac}

\newcommand{\ADM}{\mathrm{ADM}}
\newcommand{\M}{\mathcal{M}}
\newcommand{\OT}{\mathrm{OT}}

\newcommand{\E}{\mathbb{E}}

\usepackage{color}
\usepackage{pstricks}
\usepackage{ifthen}
\newrgbcolor{antoncolor}{.7 .1 .1}
\usepackage{csquotes}

\usepackage{hyperref}

\usepackage[accepted]{icml2019}

\icmltitlerunning{(q,p)-Wasserstein GANs}

\begin{document}

\twocolumn[
\icmltitle{(q,p)-Wasserstein GANs: Comparing Ground Metrics for Wasserstein GANs}

\begin{icmlauthorlist}
\icmlauthor{Anton Mallasto}{diku}
\icmlauthor{Jes Frellsen}{itu}
\icmlauthor{Wouter Boomsma}{diku}
\icmlauthor{Aasa Feragen}{diku}
\end{icmlauthorlist}

\icmlaffiliation{diku}{Department of Computer Science, University of Copenhagen, Copenhagen, Denmark}
\icmlaffiliation{itu}{Department of Computer Science, IT University of Copenhagen, Copenhagen, Denmark}

\icmlcorrespondingauthor{Anton Mallasto}{mallasto@di.ku.dk}

\vskip 0.3in
]

\printAffiliationsAndNotice{}

\begin{abstract}
Generative Adversial Networks (GANs) have made a major impact in computer vision and machine learning as generative models. Wasserstein GANs (WGANs) brought Optimal Transport (OT) theory into GANs, by minimizing the $1$-Wasserstein distance between model and data distributions as their objective function. Since then, WGANs have gained considerable interest due to their stability and theoretical framework. We contribute to the WGAN literature by introducing the family of $(q,p)$-Wasserstein GANs, which allow the use of more general $p$-Wasserstein metrics for $p\geq 1$ in the GAN learning procedure. While the method is able to incorporate any cost function as the ground metric, we focus on studying the $l^q$ metrics for $q\geq 1$. This is a notable generalization as in the WGAN literature the OT distances are commonly based on the $l^2$ ground metric. We demonstrate the effect of different $p$-Wasserstein distances in two toy examples. Furthermore, we show that the ground metric does make a difference, by comparing different $(q,p)$ pairs on the MNIST and CIFAR-10 datasets. Our experiments demonstrate that changing the ground metric and $p$ can notably improve on the common $(q,p) = (2,1)$ case.
\end{abstract}

\section{Introduction}
\emph{Generative modelling} considers learning models to generate data, such as images, text or audio. Prominent generative models include the \emph{Variational Auto-Encoders} (VAEs) \cite{kingma13} and \emph{Generative Adversial Networks} (GANs) \cite{goodfellow14}, the latter of which will be studied in this work. The generative models can be trained on unlabelled data, which is a considerable advantage over supervised models, as data labelling is expensive. The usual approach employs the \emph{manifold assumption}, stating that all meaningful data lies on a low-dimensional manifold of the sample space. Based on this assumption, one is then able to learn a map from a low dimensional distribution to the true data distribution. In this step, it is essential to quantitatively measure the discrepancy between the two distributions. To this end, one chooses a \emph{metric} or a \emph{divergence} between probability distributions. The metric should reflect modelling choices with respect to which properties of the distributions are deemed similar, or what kind of invariances one wants the metric to respect.

Traditionally, probability measures have been compared using non-metric divergence measures from information geometry, e.g.~the \emph{Kullback-Leibler (KL) divergence} and \emph{Bregman divergences}. The KL-divergence has deep connections with Bayesian statistics, where likelihood maximization in model selection can be cast as minimizing the KL-divergence.

Recently, a popular family of metrics has been provided by the theory of \emph{Optimal Transport} (OT), which studies probability distributions through a geometric framework. At its heart lie the \emph{Wasserstein metrics}, which extend the underlying metric between sample points to entire distributions. Consequently, the metrics can be used to e.g.~derive statistics between populations of probability distributions, allowing the inclusion of stochastic data objects in statistical pipelines \cite{mallasto17}. Recent algorithmic advances \cite{peyre17} have made OT widespread in the fields of machine learning and computer vision, where it has been used for e.g.~domain adaption \cite{courty17}, point embeddings \cite{muzellec18} and VAEs \cite{tolstikhin17}.

Quite notably, OT has impacted GANs. The original formulation of \citet{goodfellow14} defines GANs through a minimax game of two neural networks. One of the networks acts as a generator, whereas the other network discriminates samples based on whether they originate from the data population or not.  The minimax game results in the minimization of the \emph{Jensen-Shannon divergence} between the generated distribution and the data distribution. \citet{arjovsky17} then propose to minimize the $1$-Wasserstein distance, instead, demonstrating that the new loss function provides stability to the training. This stability was mainly attributed to the Wasserstein metric being well defined even when the two distributions do not share the same support. This results in the \emph{Wasserstein GAN} (WGAN). Other notable OT inspired variations of the original GAN are discussed below.

\subsection{Related Literature}
\label{subsec:related_lit}
The original WGAN architecture \cite{arjovsky17} enforces $k$-Lipschitz constraints through \emph{weight clipping}. An alternative to  clipping the weights is provided in \emph{Spectral Normalization GANs} (SNGANS) \cite{miyato18}, which impose Lipschitzness through $l^2$-normalization of the network weights. A body of work includes the constraints through gradient penalties, first introduced in \cite{gulrajani17}, where a penalty term for non-unit-norm gradients of the discriminator is added, resulting in the WGAN-GP. \emph{Consistency Term GANs} (CTGANs), on the other hand, penalize exceeding the Lipschitz constraint directly.

The aforementioned work focuses on training the GAN when the $1$-Wasserstein metric with the $l^2$ ground metric forms the objective function. On top of this, a body of work exists exploring the use of other OT inspired metrics and divergences. Below, we discuss some notable examples.

\citet{deshpande18} propose using the sliced Wasserstein distance \cite{bonneel15}, which computes the expectation of the Wasserstein distance between one dimensional projections of the measures. This approach allows omitting learning a discriminator, but in practice a discriminator is trained for choosing meaningful projections, essential when working with high-dimensional data. The authors report increased stability in training and show that the training objective is an upper bound for the true distance between the generator and target distribution.

\citet{genevay17}, on the other hand, rely on the favorable computational properties of relaxing the original OT problem with entropic penalization. Instead of relying on the dual Rubinstein-Kantorovich formulation, they compute the \emph{Sinkhorn divergence} \cite{cuturi13} between minibatches in the primal formulation. This also allows omitting learning a discriminator, however, the authors do propose learning a cost function, as they argue the $l^2$ ground metric is not suitable in every application. The hyperparameters of the Sinkhorn divergence allows interpolating between the $2$-Wasserstein distance and Maximum Mean Discrepancy (MMD), providing more freedom the in the metric model choice. This method also allows for a general cost function to be used, like our $(q,p)-WGAN$ method, but the experiments are limited to the $p=2$ and learned distance function cases without comparison.

\citet{wu18} introduce the \emph{Wasserstein divergence}, motivated by the gradient penalty approach on the $1$-Wasserstein metric. The divergence builds on the dual formulation, by relaxing the Lipschitz constraint. Additionally, a gradient norm penalty is included, that is considered over the support of a fixed test distribution.

\subsection{Our Contribution}
We wish to add more flexibility to WGANs by using the $p$-Wasserstein distance on top of more general $l^q$ ground metrics for $p,q\geq 1$. This is achieved through the $(q,p)$-Wasserstein GAN ($(q,p)$-WGAN), which generalizes Wasserstein GANs to allow arbitrary cost functions for the OT problem, however, we limit the scope of this paper to the $l^q$ metric case. This generalization broadens the existing WGAN literature, as mostly the $1$-Wasserstein distance with $l^2$ metric is considered.  We demonstrate the importance of the resulting flexibility in our experiements. Moreover, $(2,1)$-WGAN provides a novel way of taking into account the 1-Lipschitz constraints required in the original WGAN minimizing the 1-Wasserstein distance.

Given our $(q,p)$-WGAN implementation, we study the effect of $p$ when we fix $q=2$ in two toy examples. Additionally, we compare $p$-Wasserstein metrics based on the $l^q$ ground metric between samples for $p=1,2$ and $q=1,2$ on the MNIST and CIFAR-10 datasets. The $(q,p)$-WGANs are compared to WGAN and WGAN-GP on the CIFAR-10 dataset to assess the performance of our implementation. The experiments show, that choosing $q=1$ outperforms $q=2$ on colored image data, where as $p=2$ slightly outperforms $p=1$. Based on the results, it is clear that the metric used for GANs should be tailored to fit the needs of the application.

Finally, the OT theory suggests that the Kantorovich potentials (or discriminators) can also function as generators through their gradients. We try this on the MNIST dataset, and conclude that the generator clearly improves the results.

\section{Background}
\label{sec:background}
We briefly summarize the prequisites for this work. The methodology is founded on optimal transport, which we will revise first.  We finish the section by reviewing the mathematical details of GANs with a focus on WGANs.
\subsection{Optimal Transport}
\label{subsec:ot}
The aim in \emph{Optimal Transport} (OT) is to define a geometric framework for the study of probability measures. This is carried out by defining a \emph{cost function} between samples (e.g. the $l^2$ metric), and then studying \emph{transport plans} that relate two compared probability measures to each other while minimizing the total cost. A common example states the problem as moving a pile of dirt into another with minimal effort, by finding an optimal allocation for each grain of dirt so that the cumulative distance of dirt moved is minimized.

We start with basic definitions, and conclude by discussing the \emph{Wasserstein metric}. The interested reader may refer to \citet{villani08} for theoretical and \citet{peyre17} for computational aspects of OT.

\textbf{Optimal Transport Problem.} Let $\mu$ be a probability measure on a metric space $X$, denoted by $\mu \in \M(X)$. Let $f \colon X\to Y$ be a measurable map. Then $f_\#\mu(A):=\mu(f^{-1}(A))$ denotes the push-forward of $\mu$ with respect to $f$. Here $A$ is any measurable set in another metric space $Y$. The push-forwad can be also explained from a sampling perspective; assume $\xi$ is a random variable with distribution $\mu$. Then $f(\xi)$ has distribution $f_\#\mu$.

Given two probability measures $\mu \in \M(X)$, and $\nu \in M(Y)$, we define the set of \emph{admissable plans} by
\begin{equation}
\begin{aligned}
 &\ADM(\mu,\nu)\\
  =& \{\gamma \in \M(X\times Y) | ~(\pi_1)_\#\gamma = \mu,~(\pi_2)_\#\gamma = \nu\},
 \end{aligned}
\end{equation}
where $\pi_i$ denotes the projection onto the $i$th coordinate. In layman's terms, a joint measure on $X\times Y$ is admissable, if its marginals are $\mu$ and $\nu$.

Now, given a lower semi-continuous \emph{cost function} $c:X\times Y \to \mathbb{R}$ (such as the $l^q$ metric $d_q$), the task in optimal transport is to compute
\begin{equation}
\OT_c(\mu, \nu) := \min\limits_{\gamma \in \ADM(\mu,\nu)} \E_{\gamma}[c],
\label{def:primal_ot}
\end{equation}
where we use $\E_\mu[f]$ to denote the expectation of a function $f$ under the measure $\mu$, that is,
\begin{equation}
\E_\mu[f] = \int_X f(x)d\mu(x).
\end{equation}
Next, denote by $L^1(\mu)=\{f|~\E_\mu[f] < \infty\}$ the set of functions that have finite expectations with respect to $\mu$. Let $\varphi \in L^1(\mu)$, $\psi \in L^1(\nu)$. Then, assume $\varphi, \psi$ satisfy
\begin{equation}
\varphi(x) + \psi(y) \leq c(x,y),~\forall (x,y)\in X\times Y.
\label{def:admc}
\end{equation}
We denote the set of all such pairs by $\ADM(c)$. Then, $\OT_c(\mu,\nu)$ can be expressed in the \emph{dual formulation}
\begin{equation}
\OT_c(\mu, \nu) = \max\limits_{(\varphi, \psi)\in \ADM(c)}\left\lbrace \E_\mu[\varphi] + \E_\nu[\psi]\right\rbrace.
\label{def:dual_ot}
\end{equation}
The optimal functions $\varphi, \psi$ are called \emph{Kantorovich potentials}, and they satisfy
\begin{equation}
\varphi(x) + \psi(y) = c(x,y),~\forall (x,y)\in \mathrm{Supp}(\gamma).
\end{equation}

The Kantorovich potentials $\varphi$ and $\psi$ are intimately related. Define the $c$-transform of $\varphi$ as
\begin{equation}
\varphi^{c}:Y\to \mathbb{R},~y\mapsto \inf\limits_{x\in X} \left\lbrace c(x,y) - \varphi(x)\right\rbrace,
\label{def:c_transform}
\end{equation}
then according to the fundamental theorem of optimal transport, the Kantorovich potentials satisfy $\psi = \varphi^{c}$, and thus \eqref{def:dual_ot} can be written as
 \begin{equation}
\OT_c(\mu, \nu) = \max\limits_{(\varphi, \varphi^c) \in \ADM(c)}\left\lbrace \E_\mu[\varphi] + \E_\nu[\varphi^{c}]\right\rbrace,
 \end{equation}
 reducing the optimization to be carried out over a single function.

\textbf{Wasserstein Metric.} It turns out that the OT framework can be used to define a distance between probability distributions. Define the set 
\begin{equation}
\mathcal{P}^p_d(X)= \left\lbrace \mu \in \M(X)\left|~\int \right. d^p(x_0,x)d\mu (x) < \infty  \right\rbrace
\end{equation}
 for any $x_0\in X$. Then, $\OT_c(\mu, \nu)$ defines a metric between $\mu, \nu\in \mathcal{P}_d^p(X)$, if we choose the cost $c$ to be related to a metric $d$ on $X$, called the \emph{ground metric}, in the following way.

The $p$-Wasserstein metric $W_p$ between $\mu,\nu \in \mathcal{P}_d^p(X)$, where $(X,d)$ is a metric space, is given by 
\begin{equation}
W_p(\mu, \nu) := \left(\OT_{\sfrac{d^p}{p}}(\mu, \nu)\right)^\frac{1}{p}.
\end{equation}

When $\mu, \nu$ are absolutely continuous measures on $X=Y=\mathbb{R}^n$ with the Euclidean $l^2$ metric and $p>1$, the optimal transport plan is  induced by a unique \emph{transport map} $T\colon X\mapsto X$, for which $T_\#\mu = \nu$, given by
\begin{equation}
T = (I - \|\nabla \varphi \|^{p'-2}\nabla\varphi),
\label{eq:optimal_map}
\end{equation}
where $\varphi$ stands for the optimal Kantorovich potential in the dual formulation \eqref{def:dual_ot}, and $p^{-1} + (p')^{-1} = 1$. Therefore, computing the $p$-Wasserstein distance by the dual formulation yields us a map between the distributions, which we will later employ in the experimental section.

\textbf{The Ground Metric.} When $X=Y=\mathbb{R}^d$, commonly the $l^2$ metric is chosen as the ground metric for the $p$-Wasserstein distance. However, depending on the application, any other distance can be also considered, for example any $l^q$ distance $d_{q}$ for $q\geq 1$, given by
\begin{equation}
d_{q}(x,y) = \left(\sum_{i=1}^n |x_i - y_i|^{q}\right)^{\frac{1}{q}}.
\end{equation}
In the experimental section, we study the effect of the ground metric, when minimizing the $p$-Wasserstein distance in the context of GANs. To emphasize the ground metric, we introduce the $(q,p)$-Wasserstein distance notation
\begin{equation}
W_{q,p}(\mu, \nu) = \left(\OT_{d_q^p/p}(\mu,\nu)\right)^\frac{1}{p}.
\end{equation}
To not diverge too far from the standard notation, we assume that $q=2$ for the $p$-Wasserstein distance denoted by $W_p$.

\subsection{Generative Adversial Networks}
\emph{Generative Adversial Networks} (GANs) are a popular tool for learning data distributions \cite{goodfellow14}. The GAN approach consists of a competitive game between two networks, the \emph{generator} $g_\omega$ and the \emph{discriminator} $\varphi_{\omega'}$, with parameters $\omega$ and $\omega'$, respectively. Given the target distribution $\mu_t$ of the data, and a low-dimensional \emph{source distribution} $\mu_s$, the GAN minimax objective is given by
\begin{equation}
\begin{aligned}
\min\limits_{\omega}\max\limits_{\omega'} & \left\lbrace \E_{x\sim \mu_t}\left[\log (\varphi_{\omega'}(x))\right] \right.\\
 &+ \left.\E_{z\sim\mu_s}\left[\log(1 - \varphi_{\omega'}(g_\omega(z)))\right]\right\rbrace.
 \label{def:gan_obj}
\end{aligned}
\end{equation}
At optimality, this corresponds to minimizing the Jensen-Shannon divergence between $\mu_t$ and $(g_{\omega})_\# \mu_s$, the push-forward of the source with respect to the generator. The discriminator has range $[0,1]$, expressing the probability of a sample being from the original data distribution.

The \emph{Wasserstein GAN} introduced by \citet{arjovsky17} minimizes the $1$-Wasserstein metric instead. The authors argue that divergences such as Jensen-Shannon, or Kullback-Leibler do not behave well with respect to the generator's parameters. This is due to these divergences being singular when the two distributions do not share the same support. They then demonstrate, that the $1$-Wasserstein distance behaves in a more continuous way, and provides a meaningful loss, whose decrease corresponds to increased image quality when generating images.

Writing the $1$-Wasserstein metric in the dual form, and using that for the $(q,p)=(2,1)$ case $\varphi_{\omega'}^c = - \varphi_{\omega'}$, and $(\varphi_{\omega'}, \varphi_{\omega'}^c)\in \ADM(c) $ implies that $\varphi_{\omega'}$ is $1$-Lipschitz, the minimax objective for WGANs is written as
\begin{equation}
\min\limits_{\omega}\max\limits_{\omega'}\left\lbrace \E_{x\sim \mu_t}\left[\varphi_{\omega'}(x)\right] - \E_{z\sim \mu_s}\left[\varphi_{\omega'}(g_\omega(z))\right]\right\rbrace.
\label{def:wgan_obj}
\end{equation}

This time $\varphi_{\omega'}$ is called the \emph{critic} and not the discriminator, as its range is not limited. However, in this paper, we use either name interchangeably, or might also use the name Kantorovich potential.

 In the original paper \cite{arjovsky17}, the Lipschitz constraints are enforced through weight-clipping. This, however, only quarantees $k$-Lipschitzness for some $k$, and thus a scalar multiple of the $1$-Wasserstein distance is computed. Remarking that a function is $1$-Lipschitz if and only if its gradient has norm at most $1$ everywhere, a gradient norm penalty was introduced in the WGAN-GP method of \citet{gulrajani17}. See Subsec. \ref{subsec:related_lit} for more discussion on imposing the constraints.

\section{$(q,p)$-Wasserstein GAN}
\label{sec:method}
\begin{algorithm}[tb]
   \caption{$(q,p)$-WGAN. Batch size $m=64$, learning rate $\alpha=10^{-4}$, search space $B$, and the Adam parameters $\beta_0 = 0.5$ and $\beta_1 = 0.999$. }
   \label{alg:example}
\begin{algorithmic}
	\FOR{$\mathrm{iter}=1,...,N_\mathrm{Iterations}$}
	\STATE Sample from target $x_i\sim \mu_t$ and source $z_i\sim \mu_s$, $i=1,2,...,m$, where $m$ is the batch-size. Denote $B_x=\{x_i\}_{i=1}^m$. 
	\STATE $y_i\leftarrow g_\omega(z_i)$, denote $B_y=\{y_i\}_{i=1}^m$.
	\FOR{$t=1,...,N_{\mathrm{critic}}$}
	\STATE \emph{$\#$Define $\psi_{\omega'}$:}
	\STATE $\psi_{\omega'}(y) \leftarrow \min\limits_{x\in B}\left\lbrace\frac{1}{p}d_q^p(y,x) -\varphi_{\omega'}(x)\right\rbrace$ .
	\STATE \emph{$\#$Compute penalties:}
	\STATE $P_1 = \frac{1}{m^2}\sum_{i,j=1}^m\xi(x_i,y_j)^2$
	\STATE $P_2 = \frac{1}{4m^2}\sum_{x,y\in B_x\cup B_y}\xi(x,y)^2$
	\STATE \emph{$\#$Compute objective:}
	\STATE $L \leftarrow \frac{1}{m} \sum_{i=1}^{m}\left(\varphi_{\omega'}(x_i) + \psi_{\omega'}(y_i\right)) -P_1 - P_2$.
	\STATE \emph{$\#$Update critic:}
	\STATE $\omega' \leftarrow \omega' + \mathrm{Adam}(\nabla_{\omega'}L, \alpha, \beta_0, \beta_1)$.
	\ENDFOR 
	\STATE \emph{$\#$Compute Wasserstein loss:}
	\STATE $\leftarrow \frac{1}{m} \sum_{i=1}^{m}\left(\varphi_{\omega'}(x_i) + \psi_{\omega'}(y_i\right))$
	\STATE \emph{$\#$Update generator:}
	\STATE $\omega \leftarrow \omega - \mathrm{Adam}(\nabla_{\omega}W, \alpha, \beta_0, \beta_1)$.
	\ENDFOR
\end{algorithmic}
\label{alg:pwgan}
\end{algorithm}

\begin{figure*}[htb!]
\centering
\includegraphics[width = \linewidth]{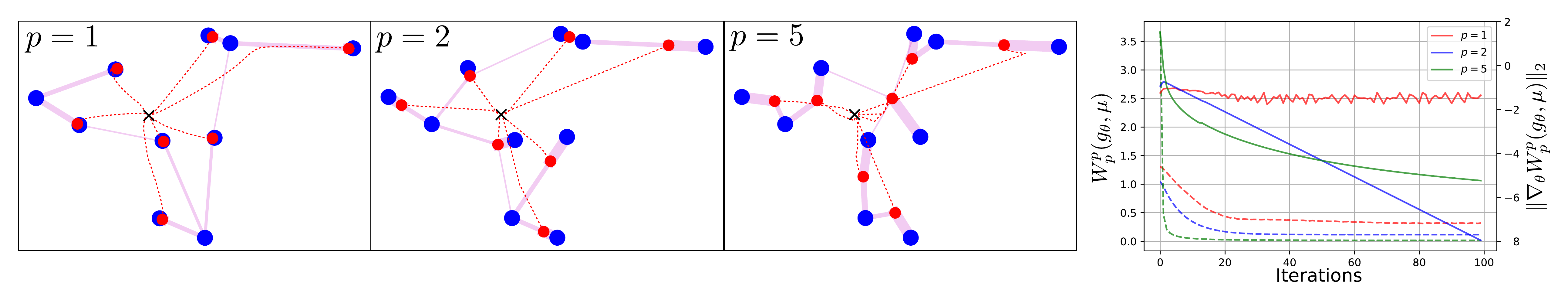}
\vspace{-0.35in}
\caption{Comparing the $p$-Wasserstein metrics for $p=1,2,5$ with the Euclidean metric. We minimize $W_p^p$ between a discrete target distribution $\mu$ with $10$ atoms (blue), by optimizing the support of a model distribution $g_\theta$ with $7$ atoms (in red). The trail of the changing support is drawn in dashed red, starting from the origo (cross). The magenta lines express the optimal mass transport between the two measures after optimization. Both measures have uniform weights. Plot on the right shows convergence for each case in distance $W_p^p$ (dashed) and gradient norm $\|\nabla_\theta W_p^p(g_\theta, \mu)\|_2$ (solid).}
\label{fig:toy_discrete}
\end{figure*}

We will now introduce the novel $(q,p)$-Wasserstein GAN ($(q,p)$-WGAN) architecture, which minimizes the $(q,p)$-Wasserstein distance between the target distribution $\mu_t$ and the approximation $(g_\omega)_\#\mu_s$. That is, the cost function is given by $c=d_q^p/p$. The objective reads
\begin{equation}
\begin{aligned}
&\min\limits_\omega W_{q,p}^p((g_\omega)_\#\mu_s, \mu_t)\\
=&\min\limits_\omega \max\limits_{(\varphi_{\omega'}, \varphi_{\omega'}^c)\in \ADM(c)} \left\lbrace\E_{x\sim \mu_t}[\varphi_{\omega'}(x)]  \right. \\
&+\left. \E_{z\sim\mu_s}[\varphi_{\omega'}^c(g_\omega(z))]\right\rbrace.
\end{aligned}
\label{def:pwgan_obj}
\end{equation}
This formulation requires one to approximate the $c$-transform defined in \eqref{def:c_transform} and to enforce the constraint $(\varphi_{\omega'}, \varphi_{\omega'}^c)\in \ADM(c)$.

\textbf{The $c$-transform.} For computing the $c$-transform, we choose a \emph{search space} $B$ for the minimization. For example, the learning procedure of the GAN is carried out through mini-batches. Hence, we can compute the discrete $c$-transform over the mini-batches. That is, given sets of samples $B_x=\{x_i\}_{i=1}^m$ and $B_y=\{y_i\}_{i=1}^m$, from the target $\mu_t$ and generator $(g_\omega)_\#\mu_s$, respectively, we compute the approximation $\varphi_{\omega'}^c$ over $B=B_x \cup B_y$
\begin{equation}
\varphi_{\omega'}^c(y_j) \approx \min\limits_{x\in B} \left\lbrace c(x, y_j) - \varphi_{\omega'}(x) \right\rbrace.
\end{equation}
In the experiments, we use both $B = B_x$ and $B = B_x \cup B_y$.

\textbf{Enforcing the constraints.} Define 
\begin{equation}
\xi(x,y) = c(x,y)-\varphi_{\omega'}(x)-\varphi^c_{\omega'}(y).
\end{equation}
Then, when training the discriminator, we add two penalty terms given by
\begin{equation}
\begin{aligned}
P_1(\varphi) &= \lambda_1\sum_{i,j=1}^m \xi(x_i,y_j)^2,\\
P_2(\varphi) &=\lambda_2\sum_{x,y\in B_x \cup B_y} \min(\xi(x,y), 0)^2.
\end{aligned}
\end{equation}
Here $P_2$ enforces $(\varphi, \varphi^c)\in \ADM(c)$ over all elements in $B_x \cup B_y$, and $P_1$ encourages pairs $(x_i,y_j)$ to belong in the support of the optimal plan.

The $(q,p)$-WGAN method is summarized in Algorithm \ref{alg:pwgan}.

\section{Comparison of $p$-Wasserstein Metrics}
\label{sec:metric_comparison}
To give some intuition about the differences between different $p$-Wasserstein metrics $W_p$, we compare the behavior of $W_p$ for $p=1,2,5$ on two toy examples. The first example consists of approximating a discrete probability measure with another discrete measure with smaller support. This example is intended to give general intuition of the behavior of the $p$-Wasserstein distance when compromises are required, however, the intuition might not translate directly into the GAN setting. The second example demonstrates fitting a $(2,p)$-WGAN to a $2$-dimensional Gaussian mixture. We abbreviate $(2,p)$-WGAN as $p$-WGAN.

\begin{figure}[htb!]
\centering
\includegraphics[width = 0.8\linewidth]{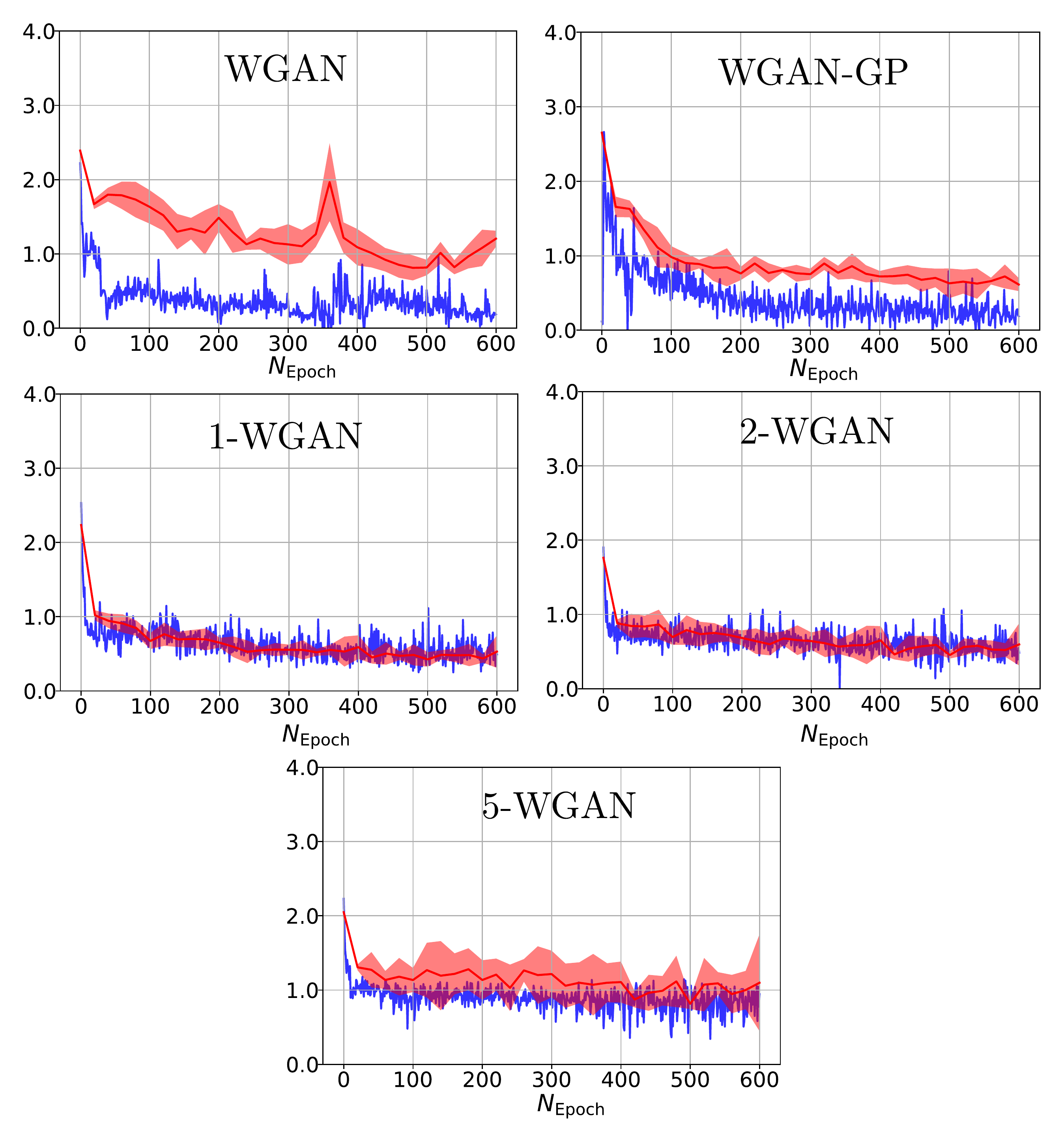}
\caption{Convergence of model distributions for the Gaussian mixture model. Objective function (approximation of the $p$-Wasserstein distance) after each epoch (for WGAN, this has been renormalized with the estimated Lipschitz constant) in blue. The true $p$-Wasserstein distance computed between the data set and the same amount of generator samples in red.}
\label{fig:toy_gan_dists}
\end{figure}

\begin{figure*}[htb!]
\centering
\includegraphics[width = 0.8\linewidth]{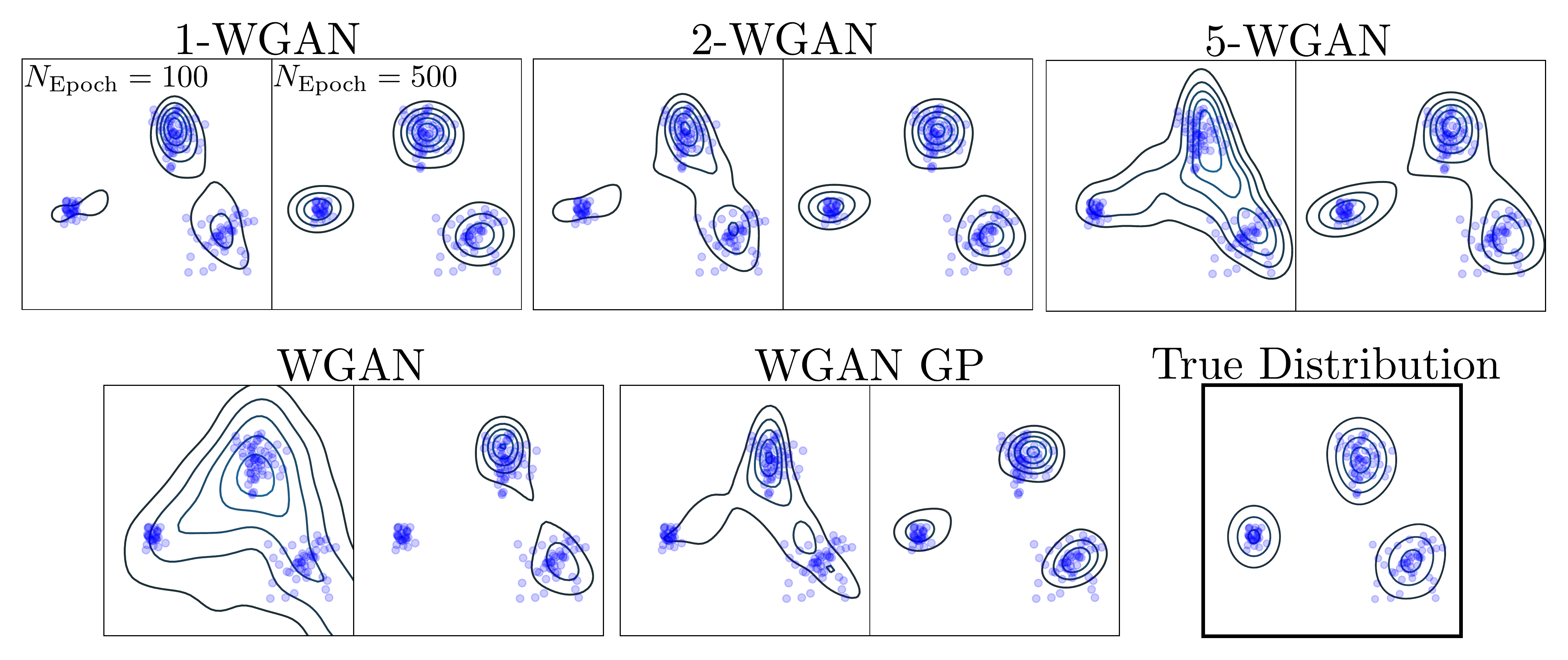}
\caption{Approximating a Gaussian mixture distribution (samples in blue) with different WGAN architectures. Presented are the results after $100$ and $500$ epochs for $(2,p)$-WGAN, abbreviated $p$-WGAN, for $p=1,2,5$. Furthermore, we present the results for the original WGAN and WGAN-GP.}
\label{fig:toy_gan_dens}
\end{figure*}
\textbf{In the first example}, the target distribution $\mu$ has $10$ atoms with uniform weights. We approximate the target with a model distribution $\nu$ with $7$ atoms and uniform weights. This objective is closely related to $k$-means clustering \cite{pollard82, canas12}. In fact, the objective would be equivalent to $k$-means, if each model distribution atom was assigned the mass of the corresponding cluster of target distribution atoms.

In Fig. \ref{fig:toy_discrete}, it is clearly seen that in the $p=1$ case, the model distribution prefers to have a support that overlaps with the target. When $p=2$, the model prefers cluster means as its support, and thus samples from the model are not exactly the same as the real samples of the target. Looking at the $p=5$ case, it seems that the model starts shrinking to the interior of the convex hull of the target's support, reducing the variance of the model distribution. Higher $p$-value seems to imply faster and more stable optimization, however, we do not witness this in the second example below (Fig. \ref{fig:toy_gan_dists}).

\textbf{In the second example}, we model a Gaussian mixture model with three clusters (cluster sizes $60$, $30$ and $50$) using a GAN that minimizes $W_p^p$. The critic and generator are Multi-Layer Perceptrons (MLPs) with ReLU activations (the output is without activation) and two fully connected hidden layers of size $128$. In addition to comparing the $p$-Wasserstein distances for $p=1,2,5$, we also compare the results to WGAN and WGAN-GP architectures.

In Fig. \ref{fig:toy_gan_dens}, the learned distributions are visualized after $100$ and $500$ epochs under the original dataset. When comparing to the true distribution, $1$-WGAN, $2$-WGAN and WGAN-GP seem to converge the fastest and provide qualitatively the best results. 5-WGAN seems to fail separating the clusters from each other, whereas WGAN expresses mode collapse. 

The convergence of each model is demonstrated in Fig. \ref{fig:toy_gan_dists}, where the objective function value and $p$-Wasserstein distance between the original dataset and the same amount of generator samples are visualized. For $1$-WGAN and $2$-WGAN, the objective function approximates well the real $p$-Wasserstein distance, whereas $5$-WGAN is more unstable. Note that in the WGAN case, the Lipschitz constant is estimated to normalize the objective function for an approximation of the $1$-Wasserstein distance. WGAN convergence is clearly more unstable than the others.

\textbf{Conclusion.} From the toy examples it is obvious, that different $p$ values result in differently behaving optimization problems. If the model is given extreme freedom (but still limited expressive power), as in the first example on discrete probability measures, higher $p$-values result in stabler optimization, but also reduces the variance. On the other hand, $p=1$ overfits by trying to overlap with the target distribution. However, this does not directly translate to the GAN example, which might be because of the model being expressive enough to match the data distribution well. Nevertheless, this example demonstrates that the $(q,p)$-WGAN models the objective Wasserstein distance well.

\section{Experiments}
\begin{figure}[htb!]
\centering
\includegraphics[width=\linewidth]{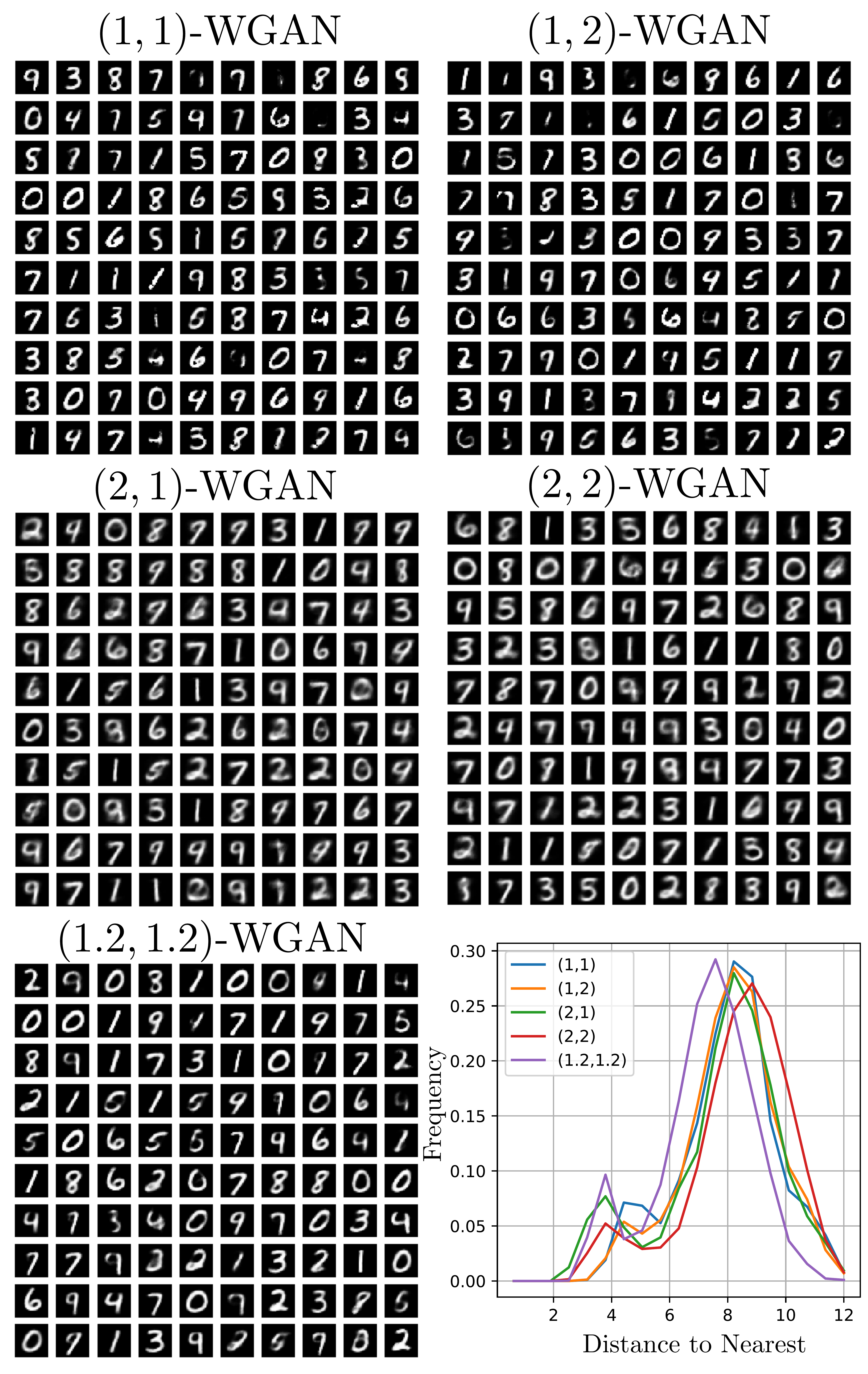}
\caption{Generated samples from different $(q,p)$-WGANS trained on the MNIST training set. Furthermore, plotted is the distribution of $l^2$ distances to closest training points of 5000 generated samples from each model.}
\label{fig:MNIST}
\end{figure}

\begin{figure*}[htb!]
\centering
\includegraphics[width=0.8\linewidth]{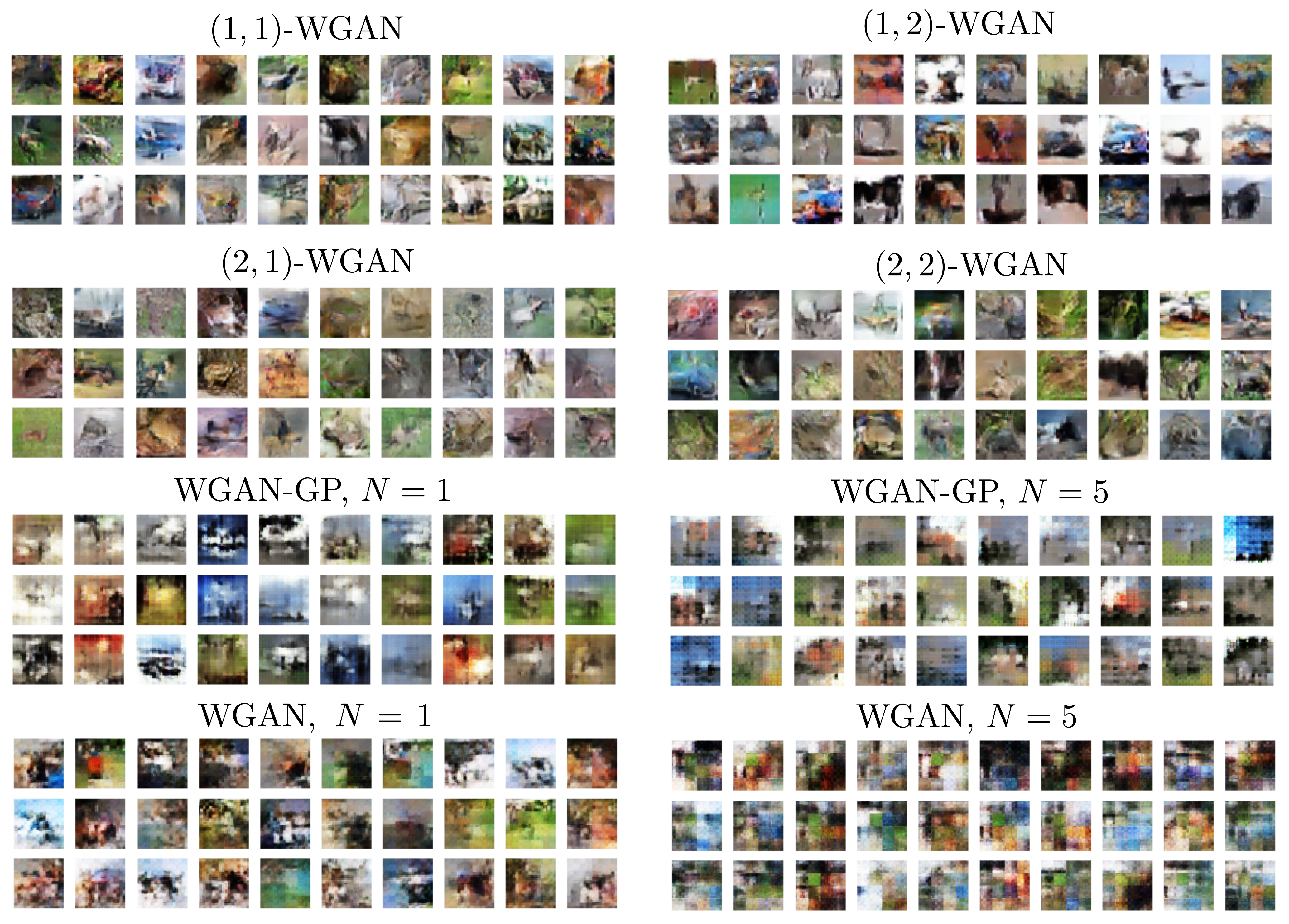}
\caption{Generated samples from different $(q,p)$-WGANS, the original WGAN and WGAN-GP trained on the CIFAR-10 training set. Here $N$ refers to the amount of critic iterations, for $(q,p)$-WGANS, this is $1$. The IS and FID scores are reported in Table \ref{table:CIFAR10}.}
\label{fig:CIFAR10}
\end{figure*}

We evaluate the performance of the $(q,p)$-WGAN method on two different datasets; MNIST \cite{mnist} and CIFAR-10 \cite{cifar}. The effect of ground metric is explored on the MNIST dataset by visually assessing the image quality. We quantify the performance of different $(q,p)$-WGANs by computing the \emph{Inception Score} (IS) \cite{is} and the \emph{Fr\'echet Inception Distance} (FID) \cite{fid} on the CIFAR-10 dataset. We use the DCGAN architecture from \cite{radford15} for CIFAR-10, and Multi-Layer Perceptrons (MLP) for MNIST, which are trained for 50K generator iterations. We use $m=64$ as the batch-size for every experiment and $N_{\mathrm{critic}}=1$ for $(q,p)$-WGANs.

\subsection{Effect of Ground Metric on MNIST.}
The MNIST dataset consists of $28\times 28$ greyscale images of hand-written digits, grouped into training and validation sets of sizes 60k and 10k, respectively. We train five different $(q,p)$-WGAN models, listed in Fig. \ref{fig:MNIST}, on the training set. We also show the distribution of distances of generated samples to closest training samples for each model, to quantify whether we are creating new digits or just memorizing the ones from the training set. Based on the first toy-example in Fig. \ref{fig:toy_discrete}, the hypothesis is that $1$-Wasserstein GAN tends to overfit to the data compared to a higher $p$ value. However, this is not evident in Fig. \ref{fig:MNIST}. 

The neural networks used are simple MLPs with 3 hidden layers (specifics in the supplementary material), that are trained for 50K generator and discriminator iterations. For the discrete $c$-transform, the search space for the minimizer is restricted to $B_x$ and $\lambda_1=\lambda_2=0$, as otherwise the model tended to collapse to single point, and $\alpha=10^{-4}$ was used as the learning rate.

The ground metric clearly affects the sharpness of produced images. When $q=2$, the generated digits have quite blurry edges. On the other hand, when $q=1$, the digits are sharp, but also more degenerate samples are produced. The sharpness can be adjusted, as shown by the samples generated by the $(1.2,1.2)$-WGAN. 

\begin{table}
\centering
\begin{tabular}{ccc}
\hline
Model&IS&FID\\
\hline
(1,1)-WGAN&$\mathbf{4.18 \pm 0.08}$&$80.7$\\
(1,2-WGAN)&$4.11\pm 0.11$&$\mathbf{78.7}$\\
(2,1)-WGAN&$3.79\pm 0.09$&$100.0$\\
(2,2)-WGAN&$4.09\pm 0.13$&$82.5$\\
WGAN, $N=1$&$2.87\pm 0.07$&$152.9$\\
WGAN, $N=5$&$2.33\pm 0.05$&$164.7$\\
WGAN-GP, $N=1$&$3.65 \pm 0.09$&$117.6$\\ 
WGAN-GP, $N=5$&$2.85 \pm 0.07$&$162.6$\\
\hline
\end{tabular}
\caption{The \emph{Inception Score} (IS) and \emph{Fr\'echet Inception Distance} (FID) for the CIFAR-10 dataset reported for four different $(q,p)$-WGANs, the original WGAN, and WGAN-GP. Here $N$ implies discriminator iterations per generator iteration. The models are trained for $50$K discriminator iterations.}.
\label{table:CIFAR10}
\end{table}

\subsection{Assessing the Quality on CIFAR-10.}
The CIFAR-10 dataset consists of $50$K $32\times 32$ color images for training. We train four different $(q,p)$-WGANs, the original WGAN, and WGAN-GP. The methods are compared by computing the IS and FID after $50$K discriminator iterations. As the original WGAN and WGAN-GP propose to use $5$ critic iterations per generator iteration, for fair comparison we carry out the training with $N_{\mathrm{critic}}=1,5$.

This time, we use the DCGAN architecture for the generator and discriminator, see supplementary for details. We use the hyperparameters proposed for WGAN and WGAN-GP by the original papers, except for the different critic iteration amounts. For $(q,p)$-WGANs, $\alpha=10^{-4}$ is the learning rate, $\lambda_1=0.1$ and $\lambda_2 = 10$, and we use $B_x\cup B_y$ as the $c$-transform search space. Restricting the search space to $B_x$ produced very blurry images.

The scores are presented in Table \ref{table:CIFAR10}, and example samples in Fig. \ref{fig:CIFAR10}. Based on Table \ref{table:CIFAR10}, the $(q,p)$-WGANs outperform WGAN and WGAN-GP. The IS and FID scores are notably higher when $q=1$. In the $q=2$ case, the $(2,2)$-Wasserstein metric scores better than the $(2,1)$-Wasserstein metric, but in the $q=1$ case the difference is marginal.

\subsection{Kantorovich Potentials as Generators}
\begin{figure}[htb!]
\centering
\includegraphics[width=1\linewidth]{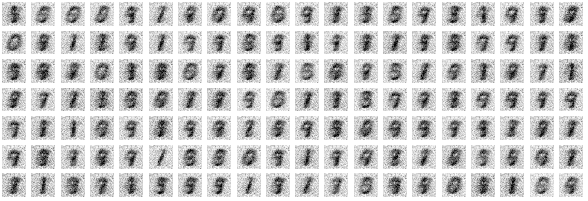}
\caption{Samples generated by \emph{only} the discriminator on the MNIST dataset.}
\label{fig:discriminator_samples}
\end{figure}

As pointed out earlier, the learned Kantorovich potentials can also be used as generators by computing the optimal transport map using \eqref{eq:optimal_map}. To see if this is applicable in practice, we train the Kantorovich potentials for the $(2,2)$-WGAN for $100$K iterations on MNIST. Although the samples clearly look like digits, we conclude that the quality of the samples in Fig. \ref{fig:discriminator_samples} is much worse than with a generator.

\section{Conclusion}
We introduced the $(q,p)$-WGAN to allow the study of different $p$-Wasserstein metrics and $l^q$ ground metrics in the GAN setting. We show that these parameters do have a definite effect on GAN training; $1$-Wasserstein metric encourages models to overfit, whereas too high $p$ causes too low variance in the model. The FID scores from the CIFAR-10 dataset indicate that $p=2$ performs better compared to $p=1$.  We also demonstrate that the $l^1$ metric outperforms $l^2$ when learning the distribution of colored images of the CIFAR-10 dataset. Moreover, the experiments show that our implementation is competitive with the literature, outperforming the WGAN and WGAN-GP implementations. 

The $(q,p)$-WGAN incorporates the $\ADM(c)$ constraints directly on the neural network modelling the Kantorovich potential $\varphi$. The other WGAN implementations, on the other hand, seem to focus on enforcing Lipschitzness and using the knowledge $\psi = \varphi^c$, which are implications of the $\ADM(d_2)$ constraints. Working with the general constraint allows for more flexibility in the modelling choices, resulting in improved performance, as we demonstrated. However, our implementation of taking the constraints into account leaves room for improvement, as we had to use considerably different hyperparameters on MNIST and CIFAR-10 to achieve stable training. We hope that our results on the importance of the ground metric and the $p$ parameter inspire research into more efficient implementations to incorporate general cost functions.

Although the generative properties of the Kantorovich potentials did not perform well in our experiment, this might be implementation dependant. We learned the Kantorovich potential field, but in some applications, learning the gradient field directly can be more fruitful \cite{chmiela17}.

Finally, from the theoretical perspective, choosing $p=2$ and a Riemannian ground metric $d$ results in a Riemannian structure over the manifold of probability measures, shown by \citet{otto01}. Thus Riemannian geometry can be used to study the probability distributions. When $p\neq 2$, a Finslerian structure is induced instead \cite{agueh12}. In layman's terms, Riemannian structure allows the study of lengths and comparison of directions through local inner-products, whereas Finslerian structures provide only direction dependant length-structures. Thus the Riemannian structure results in a more powerful framework for studying the geometry of probability distributions, and possibly GANs.

\section*{Acknowledgements}
AM and AF were supported by Centre for Stochastic Geometry and Advanced Bioimaging, funded by a grant from the Villum Foundation.

\bibliography{references}

\begin{thebibliography}{26}
\providecommand{\natexlab}[1]{#1}
\providecommand{\url}[1]{\texttt{#1}}
\expandafter\ifx\csname urlstyle\endcsname\relax
  \providecommand{\doi}[1]{doi: #1}\else
  \providecommand{\doi}{doi: \begingroup \urlstyle{rm}\Url}\fi

\bibitem[Agueh(2012)]{agueh12}
Agueh, M.
\newblock Finsler structure in the p-{W}asserstein space and gradient flows.
\newblock \emph{Comptes Rendus Mathematique}, 350\penalty0 (1-2):\penalty0
  35--40, 2012.

\bibitem[Arjovsky et~al.(2017)Arjovsky, Chintala, and Bottou]{arjovsky17}
Arjovsky, M., Chintala, S., and Bottou, L.
\newblock Wasserstein {GAN}.
\newblock \emph{arXiv preprint arXiv:1701.07875}, 2017.

\bibitem[Bonneel et~al.(2015)Bonneel, Rabin, Peyr{\'e}, and Pfister]{bonneel15}
Bonneel, N., Rabin, J., Peyr{\'e}, G., and Pfister, H.
\newblock Sliced and {R}adon {W}asserstein barycenters of measures.
\newblock \emph{Journal of Mathematical Imaging and Vision}, 51\penalty0
  (1):\penalty0 22--45, 2015.

\bibitem[Canas \& Rosasco(2012)Canas and Rosasco]{canas12}
Canas, G. and Rosasco, L.
\newblock Learning probability measures with respect to optimal transport
  metrics.
\newblock In \emph{Advances in Neural Information Processing Systems}, pp.\
  2492--2500, 2012.

\bibitem[Chmiela et~al.(2017)Chmiela, Tkatchenko, Sauceda, Poltavsky,
  Sch{\"u}tt, and M{\"u}ller]{chmiela17}
Chmiela, S., Tkatchenko, A., Sauceda, H.~E., Poltavsky, I., Sch{\"u}tt, K.~T.,
  and M{\"u}ller, K.-R.
\newblock Machine learning of accurate energy-conserving molecular force
  fields.
\newblock \emph{Science advances}, 3\penalty0 (5):\penalty0 e1603015, 2017.

\bibitem[Courty et~al.(2017)Courty, Flamary, Tuia, and Rakotomamonjy]{courty17}
Courty, N., Flamary, R., Tuia, D., and Rakotomamonjy, A.
\newblock Optimal transport for domain adaptation.
\newblock \emph{IEEE transactions on pattern analysis and machine
  intelligence}, 39\penalty0 (9):\penalty0 1853--1865, 2017.

\bibitem[Cuturi(2013)]{cuturi13}
Cuturi, M.
\newblock Sinkhorn distances: Lightspeed computation of optimal transport.
\newblock In \emph{Advances in neural information processing systems}, pp.\
  2292--2300, 2013.

\bibitem[Deshpande et~al.(2018)Deshpande, Zhang, and Schwing]{deshpande18}
Deshpande, I., Zhang, Z., and Schwing, A.
\newblock Generative modeling using the sliced {W}asserstein distance.
\newblock In \emph{Proceedings of the IEEE Conference on Computer Vision and
  Pattern Recognition}, pp.\  3483--3491, 2018.

\bibitem[Genevay et~al.(2017)Genevay, Peyr{\'e}, and Cuturi]{genevay17}
Genevay, A., Peyr{\'e}, G., and Cuturi, M.
\newblock Learning generative models with {S}inkhorn divergences.
\newblock \emph{arXiv preprint arXiv:1706.00292}, 2017.

\bibitem[Goodfellow et~al.(2014)Goodfellow, Pouget-Abadie, Mirza, Xu,
  Warde-Farley, Ozair, Courville, and Bengio]{goodfellow14}
Goodfellow, I., Pouget-Abadie, J., Mirza, M., Xu, B., Warde-Farley, D., Ozair,
  S., Courville, A., and Bengio, Y.
\newblock Generative adversarial nets.
\newblock In \emph{Advances in neural information processing systems}, pp.\
  2672--2680, 2014.

\bibitem[Gulrajani et~al.(2017)Gulrajani, Ahmed, Arjovsky, Dumoulin, and
  Courville]{gulrajani17}
Gulrajani, I., Ahmed, F., Arjovsky, M., Dumoulin, V., and Courville, A.~C.
\newblock Improved training of {W}asserstein {GAN}s.
\newblock In \emph{Advances in Neural Information Processing Systems}, pp.\
  5767--5777, 2017.

\bibitem[Heusel et~al.(2017)Heusel, Ramsauer, Unterthiner, Nessler, and
  Hochreiter]{fid}
Heusel, M., Ramsauer, H., Unterthiner, T., Nessler, B., and Hochreiter, S.
\newblock {GAN}s trained by a two time-scale update rule converge to a local
  {N}ash equilibrium.
\newblock In \emph{Advances in Neural Information Processing Systems}, pp.\
  6626--6637, 2017.

\bibitem[Kingma \& Welling(2013)Kingma and Welling]{kingma13}
Kingma, D.~P. and Welling, M.
\newblock Auto-encoding variational {B}ayes.
\newblock \emph{arXiv preprint arXiv:1312.6114}, 2013.

\bibitem[Krizhevsky \& Hinton(2009)Krizhevsky and Hinton]{cifar}
Krizhevsky, A. and Hinton, G.
\newblock Learning multiple layers of features from tiny images.
\newblock Technical report, Citeseer, 2009.

\bibitem[LeCun et~al.(1998)LeCun, Bottou, Bengio, and Haffner]{mnist}
LeCun, Y., Bottou, L., Bengio, Y., and Haffner, P.
\newblock Gradient-based learning applied to document recognition.
\newblock \emph{Proceedings of the IEEE}, 86\penalty0 (11):\penalty0
  2278--2324, 1998.

\bibitem[Mallasto \& Feragen(2017)Mallasto and Feragen]{mallasto17}
Mallasto, A. and Feragen, A.
\newblock Learning from uncertain curves: The 2-{W}asserstein metric for
  {G}aussian processes.
\newblock In \emph{Advances in Neural Information Processing Systems}, pp.\
  5660--5670, 2017.

\bibitem[Miyato et~al.(2018)Miyato, Kataoka, Koyama, and Yoshida]{miyato18}
Miyato, T., Kataoka, T., Koyama, M., and Yoshida, Y.
\newblock Spectral normalization for generative adversarial networks.
\newblock \emph{arXiv preprint arXiv:1802.05957}, 2018.

\bibitem[Muzellec \& Cuturi(2018)Muzellec and Cuturi]{muzellec18}
Muzellec, B. and Cuturi, M.
\newblock Generalizing point embeddings using the {W}asserstein space of
  elliptical distributions.
\newblock \emph{arXiv preprint arXiv:1805.07594}, 2018.

\bibitem[Otto(2001)]{otto01}
Otto, F.
\newblock The geometry of dissipative evolution equations: the porous medium
  equation.
\newblock \emph{Journal Communications in Partial Differential Equations},
  26:\penalty0 101--174, 2001.

\bibitem[Peyr{\'e} \& Cuturi(2017)Peyr{\'e} and Cuturi]{peyre17}
Peyr{\'e}, G. and Cuturi, M.
\newblock Computational optimal transport.
\newblock Technical report, 2017.

\bibitem[Pollard(1982)]{pollard82}
Pollard, D.
\newblock Quantization and the method of k-means.
\newblock \emph{IEEE Transactions on Information theory}, 28\penalty0
  (2):\penalty0 199--205, 1982.

\bibitem[Radford et~al.(2015)Radford, Metz, and Chintala]{radford15}
Radford, A., Metz, L., and Chintala, S.
\newblock Unsupervised representation learning with deep convolutional
  generative adversarial networks.
\newblock \emph{arXiv preprint arXiv:1511.06434}, 2015.

\bibitem[Salimans et~al.(2016)Salimans, Goodfellow, Zaremba, Cheung, Radford,
  and Chen]{is}
Salimans, T., Goodfellow, I., Zaremba, W., Cheung, V., Radford, A., and Chen,
  X.
\newblock Improved techniques for training {GAN}s.
\newblock In \emph{Advances in Neural Information Processing Systems}, pp.\
  2234--2242, 2016.

\bibitem[Tolstikhin et~al.(2017)Tolstikhin, Bousquet, Gelly, and
  Schoelkopf]{tolstikhin17}
Tolstikhin, I., Bousquet, O., Gelly, S., and Schoelkopf, B.
\newblock Wasserstein auto-encoders.
\newblock \emph{arXiv preprint arXiv:1711.01558}, 2017.

\bibitem[Villani(2008)]{villani08}
Villani, C.
\newblock \emph{Optimal transport: old and new}, volume 338.
\newblock Springer Science \& Business Media, 2008.

\bibitem[Wu et~al.(2018)Wu, Huang, Thoma, Acharya, and Van~Gool]{wu18}
Wu, J., Huang, Z., Thoma, J., Acharya, D., and Van~Gool, L.
\newblock Wasserstein divergence for gans.
\newblock In \emph{Computer Vision -- ECCV 2018}, pp.\  673--688, Cham, 2018.
  Springer International Publishing.
\newblock ISBN 978-3-030-01228-1.

\end{thebibliography}
\bibliographystyle{icml2019}

\clearpage
\section*{Supplementary Material}
We present below the two neural network architectures used in the paper, one Multi-layer Perceptron, and one Convolutional Neural Network based on the DCGAN architecture.
\subsection*{MNIST}
\begin{table}[htb!]
\centering
\begin{tabular}{cc}
\textbf{Discriminator} $\varphi$ & \textbf{Generator} $g$ \\
\hline
Input: $28\times 28$-vectors & Input: 128-dimensional noise\\
\hline
Linear($28\times28$, 1024), LeakyReLU(0.2), Dropout(0.3) & Linear(128, $2\times 128$), LeakyReLU(0.2)\\
Linear($8\times 128$, $4\times 128$), LeakyReLU(0.2), Dropout(0.3) & Linear($2\times 128$, $4\times 128$), LeakyReLU(0.2)\\
Linear($4\times 128$, $2\times 128$), LeakyReLU(0.2), Dropout(0.3) & Linear($4\times 128$, $8\times 128$), Tanh\\
Linear($2\times 128$, 1)& Linear($8\times 128$, $28\times 28$), LeakyReLU(0.2)\\
\hline
\end{tabular}
\caption{Discriminator and Generator architectures for the MNIST experiment}
\end{table}

\subsection*{CIFAR-10}
\begin{table}[htb!]
\centering
\begin{tabular}{cc}
\textbf{Discriminator} $\varphi$ & \textbf{Generator} $g$ \\
\hline
Input: $28*28$-vectors & Input: 128-dimensional noise\\
\hline
Conv(3, 128), LeakyReLU(0.2) & Linear(128, $4\times 4\times 4\times 128$),  ReLU, Reshape($4\times 128, 4,4$) \\
Conv(128, $2\times 128$), LeakyReLU(0.2) & Deconv($4\times 128$, $2\times 128$), BatchNorm($2\times 128$), ReLU\\
Conv($2\times 128$, $4\times 128$), LeakyReLU(0.2) & Deconv($2\times 128$, 128), BatchNorm(128), ReLU\\
Linear($4\times 4 \times 4 \times 128$, 1)& Deconv(128, 3), Tanh\\
\hline
\end{tabular}
\caption{Discriminator and Generator architectures for the CIFAR-10 experiment. For Deconv, kernel size is 2 and stride 2 and padding 0. For Conv, kernel size is 3, stride 2 and padding 1.}
\end{table}
\end{document}